\documentclass[letterpaper,journal]{IEEEtran}
\usepackage{amsmath,amsfonts}
\usepackage{algorithmic}
\usepackage{array}
\usepackage[caption=false,font=normalsize,labelfont=sf,textfont=sf]{subfig}
\usepackage{textcomp}
\usepackage{stfloats}
\usepackage{url}
\usepackage{verbatim}
\usepackage{graphicx}
\def\BibTeX{{\rm B\kern-.05em{\sc i\kern-.025em b}\kern-.08em
    T\kern-.1667em\lower.7ex\hbox{E}\kern-.125emX}}
\usepackage{balance}
\usepackage{bm}
\usepackage[colorlinks=true,    
  linkcolor=red,                
  citecolor=blue,               
  urlcolor=magenta,             
  bookmarksnumbered]{hyperref}  
\usepackage{lettrine}
\usepackage{microtype} 
\usepackage{multirow}
\usepackage[table,xcdraw]{xcolor}
\usepackage{pifont} 
\usepackage{bbding} 
\usepackage{fontawesome} 
\usepackage{float}
\usepackage[normalem]{ulem}
\useunder{\uline}{\ul}{}

\begin{document}
\title{Few-Shot Generative Model Adaption via Identity Injection and Preservation}
\author{Yeqi He, Liang Li, Jiehua Zhang, Yaoqi Sun, Xichun Sheng, Zhidong Zhao,~\IEEEmembership{Member, IEEE} and Chenggang Yan
\thanks{Corresponding author: Liang Li.}
\thanks{Yeqi He is with the School of Cyberspace Security, Hangzhou Dianzi University, Hangzhou, China (e-mail:yeqihe@hdu.edu.cn).}
\thanks{Liang Li is with the Institute of Computing Technology, Chinese Academy
of Sciences, Beijing, China (e-mail: liang.li@ict.ac.cn)}
\thanks{Jiehua Zhang is with the School of Software Engineering, Xi'an Jiaotong University, Xi'an, China (email: jiehua.zhang@stu.xjtu.edu.cn)}
\thanks{Yaoqi Sun is with the Lishui University and Lishui Institute of Hangzhou Dianzi University, Lishui, China (e-mail: sunyq2233@163.com)}
\thanks{Xichun Sheng is with the School  of Faculty of Applied Science, Macao Polytechnic University, Macao, China(email:p2314922@mpu.edu.mo)}
\thanks{Zhidong Zhao is with the School of Cyberspace Security, Hangzhou Dianzi University, Hangzhou, China (e-mail: zhaozd@hdu.edu.cn)}
\thanks{Chenggang Yan is with the School of Communication Engineering, Hangzhou Dianzi University, Hangzhou, China (e-mail: cgyan@hdu.edu.cn)}
}

\maketitle

\begin{abstract}
    Training generative models with limited data presents severe challenges of mode collapse.  
    A common approach is to adapt a large pretrained generative model upon a target domain with very few samples (fewer than 10), known as few-shot generative model adaptation.
    However, existing methods often suffer from forgetting source domain identity knowledge during adaptation, which degrades the quality of generated images in the target domain. 
    To address this, we propose \textbf{I}dentity \textbf{I}njection and \textbf{P}reservation (\textbf{I\textsuperscript{2}P}), which leverages identity injection and consistency alignment to preserve the source identity knowledge.
    Specifically, we first introduce an identity injection module that integrates source domain identity knowledge into the target domain's latent space, ensuring the generated images retain key identity knowledge of the source domain.
    Second, we design an identity substitution module, which includes a style-content decoupler and a reconstruction modulator, to further enhance source domain identity preservation.
    We enforce identity consistency constraints by aligning features from identity substitution, thereby preserving identity knowledge.
    Both quantitative and qualitative experiments show that our method achieves substantial improvements over state-of-the-art methods on multiple public datasets and 5 metrics.
\end{abstract}

\section{Introduction}
\label{sec1:intro}

\lettrine{G}{enerative} models have achieved remarkable success in various domains, such as image generation~\cite{goodfellow2020generative, karrasStyleBasedGeneratorArchitecture2019g, galStyleGANNADACLIPGuidedDomain2021c,10154005}, image-to-image translation~\cite{10159430,10399932}, and image inpainting~\cite{yeh2017semantic,9113276}, by mapping latent codes extracted from a standard distribution to vivid images. 
Nevertheless, the training of generative models relies heavily on large amounts of high-quality training data and significant computational resources.
In practice, insufficient training data can lead to overfitting and mode collapse, negatively affecting the diversity and quality of the generated images.

\begin{figure}[t]
\centering
\includegraphics[width=1\columnwidth]{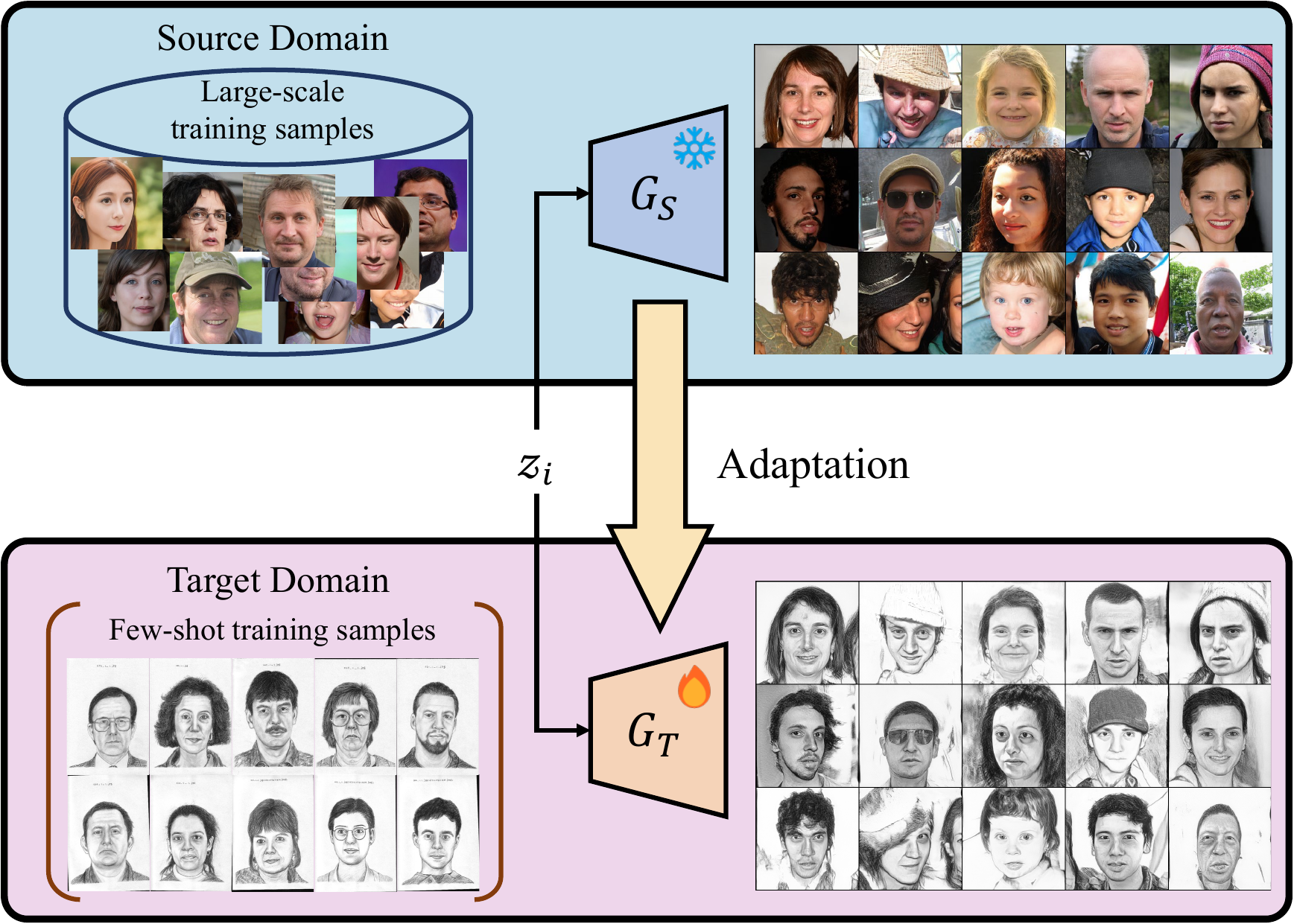} 
\caption{Few-shot generative model. Given a source generative model $G_S$ trained on a large-scale training dataset (such as FFHQ), and adapt to the target domain to get the target generative model $G_T$ by using extremely few (such as 10) training datasets}
\label{fig:first}
\end{figure}

To overcome the data scarcity challenge, researchers typically adapt pre-trained source models to target domains through three principal approaches: parameter fine-tuning~\cite{wang2018transferring, bartunov2018few}, regularization-based optimization~\cite{li2020few, XIAO2025106946}, and network perturbation~\cite{noguchi2019image, wang2020minegan}. 
As depicted in Fig.~\ref{fig:first}, these methods aim to preserve source domain identity while achieving style transfer. 
Although effective with moderate target data (e.g., 100 samples), they exhibit critical limitations under extreme few-shot conditions (e.g., 10 samples) to alleviate severe overfitting manifested through training set artifact replication, and identity degradation with mode collapse, where diminished diversity undermines both image quality and cross-domain generalization capabilities. 
This dual failure mode stems from the fundamental tension between style adaptation and identity preservation in low-data regimes.

Current approaches mainly adopt two strategies: kernel modulation~\cite{pan2024few,zhaoFewshotImageGeneration2023c,kim2022dynagan,zhao2023exploring} that selectively updates less critical generator kernels while freezing key parameters, and model regularization~\cite{ojhaFewshotImageGeneration2021c,xiao2022few,yang2023one,zhao2022closer,10483074_csvt} that aligns cross-domain distributions. 
However, both struggle with the intrinsic style-content entanglement dilemma. 
Kernel modulation methods usually fail to preserve source domain identity due to imprecise kernel importance estimation, as they lack explicit mechanisms to distinguish content-related kernels from style-specific ones. 
Model regularization methods inadvertently constrain style diversity by enforcing domain-level similarity, while maintaining structural consistency through distribution alignment. 
This fundamental conflict between precise style transfer and robust identity preservation leads to artifacts in generated images, where either the domain-specific styles become oversmoothed or the source identity characteristics get distorted. 
Thus, existing methods remain inadequate for achieving an optimal balance between adaptation fidelity and output diversity.

In this paper, we introduce the \textbf{I}dentity \textbf{I}njection and \textbf{P}reservation~(\textbf{I\textsuperscript{2}P}) algorithm for few-shot generative model adaptation, which injects the source identity knowledge and then decouples and modulates image features to integrate and retain pre-trained source domain identity knowledge.

To retain the source domain identity in the generated images, we first propose the identity injection module as a disentanglement-aware adaptation mechanism.
Inspired by adaptive instance normalization techniques~\cite{huang2017arbitrary} and latent space noise fusion techniques~\cite{chung2024style}, we extracts and integrates latent features $\mathbf{w}_S^i$ and $\mathbf{w}_T^i$ from both the source and target domain generators' latent spaces. 
The integrated features are then injected into the target domain generator's latent space, guiding the target domain's mapping network to learn and retain source domain identity knowledge.
This effectively incorporates source domain identity into the target domain generator's training, enhancing identity consistency between domains and mitigating identity drift.

Building upon this foundation, we design the identity substitution module equipped with comprehensive identity consistency constraints, which include content constraint $\mathcal{L}_{c}$, style constraint $\mathcal{L}_{s}$, and synthesis constraint $\mathcal{L}_{r}$. 
The process begins with extracting deep features from raw training images and both source and target domain generated images by using CLIP~\cite{rameshHierarchicalTextConditionalImage2022t}. 
We introduce a specialized style-content decoupler then disentangles the corresponding style features $\bm{S}$ and content features $\bm{C}$.
Next, we construct the content constraint and the style constraint to preserve the identity knowledge of the source domain and adapt the style of the training set, respectively. 
Furthermore, to ensure the integrity of decomposed representations, we establish synthesis constraints through a reconstruction modulator that explicitly governs the fusion process between style features $\bm{S}$ and content features $\bm{C}$ to synthesize modulated features $\bm{M}$. This closed-loop control mechanism effectively prevents identity-style entanglement during constraint optimization while enabling coherent knowledge recombination.

The main contributions are summarized as follows:

\begin{itemize}
\item We propose I\textsuperscript{2}P, a method that includes identity injection, identity substitution, and identity consistency, which can efficiently adapt style transfer while preserving the identity knowledge of the source domain.
\item We propose the identity injection module that injects identity knowledge from the source domain into the target domain, enhancing consistency and preventing identity drift.
\item We propose the identity preservation based on the identity substitution module with the identity consistency constraints, effectively aligning cross-domain identity and style for identity preservation and style transfer.
\item Compared with the state-of-the-art methods, our method shows promising results in both quantitative and qualitative experiments.
\end{itemize}
\section{Related Work}
\label{sec2:related}

\subsection{Few-shot image generation}
\label{sec2:few-shot-IG}


Few-shot image generation addresses the challenge of adapting pre-trained generative models to unseen domains with severely limited data (typically 10 images). While direct fine-tuning of GANs~\cite{goodfellow2020generative, bartunov2018few, wang2020minegan} provides baseline solutions, these methods often suffer from overfitting and mode collapse due to insufficient data diversity. Recent advances tackle this issue through geometric, statistical, and parameter-space regularization strategies. The CDC~\cite{ojhaFewshotImageGeneration2021c} pioneers geometric consistency by enforcing pairwise instance distance preservation between source and target domains through contrastive learning. Extending geometric constraints, RSSA~\cite{xiao2022few} introduces a dual-alignment framework that coordinates spatial structure coherence in GAN inversion latent space with adversarial distribution matching, effectively decoupling structural and stylistic adaptation. Concurrently, parameter-space optimization approaches have gained traction: AdAM~\cite{zhaoFewshotImageGeneration2023c} leverages Fisher information to identify and freeze domain-invariant generator kernels, enabling selective fine-tuning of style-related parameters. Building on parameter importance analysis, SGP~\cite{pan2024few} employs optimal transport theory to mathematically formalize knowledge preservation through Wasserstein-distance-based kernel modulation, achieving theoretically grounded balance between domain adaptation and source knowledge retention. These methods collectively demonstrate that structured regularization in either feature space or parameter space is critical for stabilizing few-shot generation, though existing methods remain inadequate for achieving an optimal balance between adaptation fidelity and output diversity.

\subsection{Decomposition}
\label{sec2:decomposition}


In the field of disentangled representation learning, a common approach is to decompose a model into two components: one that captures domain-specific variations and another that encodes domain-agnostic knowledge. Khosla et al.~\cite{khosla2012undoing} pioneered this paradigm by demonstrating that support vector machines (SVMs) can be decomposed into domain-specific bias parameters and domain-agnostic weight vectors. Their analysis revealed that retaining only the invariant weights during cross-domain inference statistically reduces distributional divergence by constraining the hypothesis space. This concept has been systematically extended to neural networks by Li et al.~\cite{li2017deeper}, who proposed layer-wise disentanglement through adaptive parameter freezing, where domain-specific neurons are dynamically deactivated based on feature activation patterns.
Recent architectural innovations further refine this decomposition principle. For instance, Chattopadhyay et al.~\cite{chattopadhyay2020learning} designed domain-specific binary masks applied to final feature representations, enforcing sparsity-driven separation through entropy constraints that theoretically minimize mutual information between domain-specific and invariant features. Concurrently, Piratla et al.~\cite{piratla2020efficient} explored low-rank weight matrix decomposition, identifying that domain-agnostic features predominantly reside in the principal subspace of network parameters, while domain-specific variations occupy residual orthogonal components. These methodologies collectively demonstrate that explicit model decomposition, when combined with geometric constraints on parameter spaces, enhances generalization beyond empirical risk minimization frameworks.
\section{Method}
\label{sec3:method}

\begin{figure*}[t]
\centering
\includegraphics[width=1\textwidth]{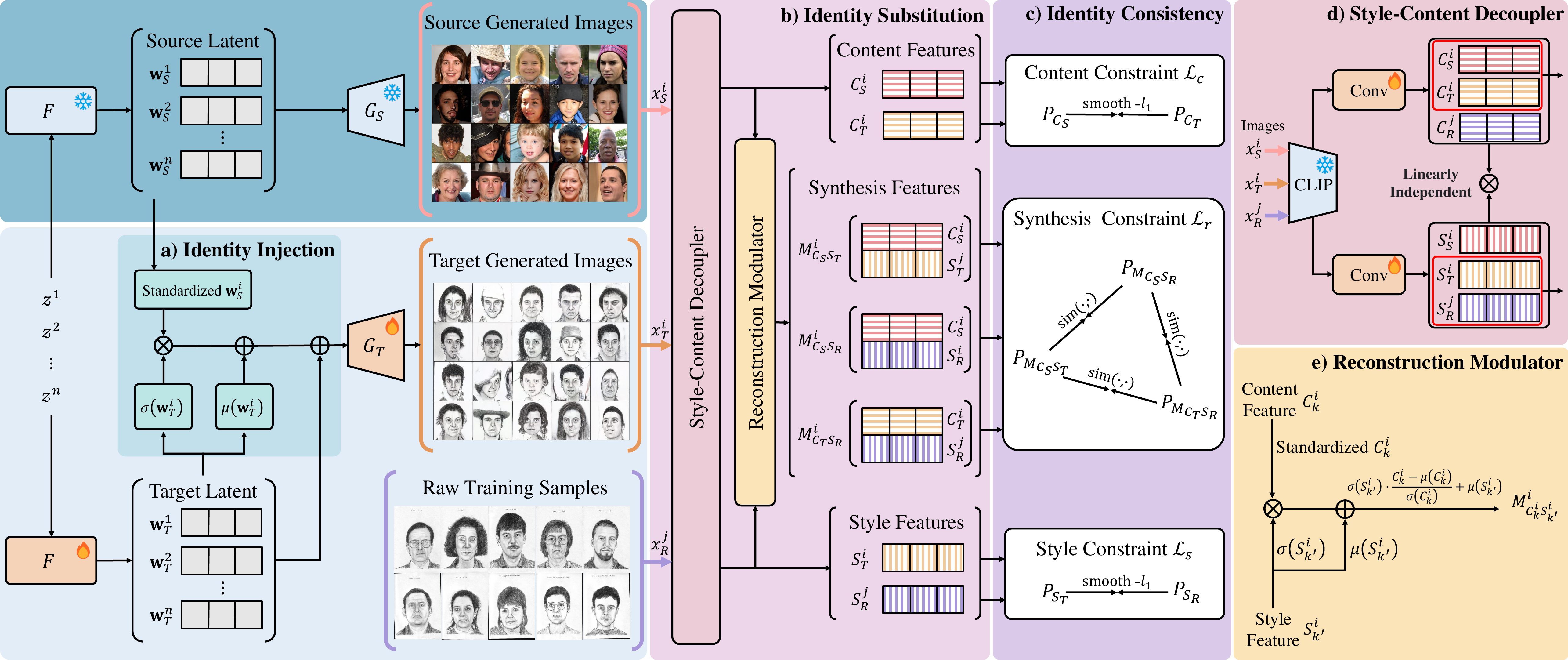}
\caption{The framework of our method. 
a) Identity injection obtains and fuses the latent features of the mapping networks in the source and target domains, thereby guiding the target domain mapping network to remain the source domain identity knowledge.
b) Identity substitution module includes a style-content decoupler that decomposes style information features and identity knowledge features, and a reconstruction modulator for reconstructing features using style information features and identity knowledge features. 
c) Identity consistency consist of style constraint $\mathcal{L}_s$, content constraint $\mathcal{L}_c$, and synthesis constraint $\mathcal{L}_r$. 
d) Internal structure of style-content decoupler. 
e) Internals of the reconstruction modulator.}
\label{fig:main}
\end{figure*}

In this section, we initially review the formulation of few-shot generative model adaptation and provide an overview of the proposed method in the Sec.~\ref{sec3:overview}. Subsequently, the Sec.~\ref{sec3:injection} offers a detailed description of identity injection. Finally, the Sec.~\ref{sec3:substitution} and Sec.~\ref{sec3:consistency} subsections address the key components of identity preservation, namely identity substitution and identity consistency.

\subsection{Overview and framework}
\label{sec3:overview}

Given a generator $G_S$ pre-trained on source domain $\mathcal{D}_S$, it maps a noise vector $z\sim p_z(z)\in Z$ to an image $G_S(z)$ in the pixel space. 
The objective of few-shot generative model adaptation is to adapt $G_S$ from the source domain to the target domain, resulting in a target generator $G_T$, using only a few samples from the target domain $\mathcal{D}_T$.
The standard fine-tuning approach involves initializing $G_T$ with $G_S$ and subsequently fine-tuning $G_T$ on the target domain dataset $\mathcal{D}_T$ through an adversarial training process. 
The optimization objective is formulated as follows:

\begin{equation}
\begin{aligned}
    \mathcal{L}_{adv}=&\mathbb{E}_{x\sim \mathcal{D}_T}[\log(1-D(x))]\\+&\mathbb{E}_{z\sim p(z)}[\log(D(G_T(z)))]
\end{aligned}
\end{equation}
where $D$ represents a learnable discriminator.

Most methods, including kernel modulation and model regularization, usually suffer from a loss of identity knowledge and reduced diversity in generated images. 
To mitigate these issues, we propose Identity Injection and Preservation (I\textsuperscript{2}P), which encompasses three key components: identity injection, identity substitution, and identity consistency. 
Firstly, we introduce an identity injection module in the latent space, as illustrated in Fig.~\ref{fig:main} a), which injects content information from the source domain into the latent space of the target domain.
This guides the mapping network of the target domain to learn and generate a latent feature for the target domain while retaining source domain identity knowledge.
Secondly, we employ a pair of modules for identity substitution: the style-content decoupler and the reconstruction modulator. 
These modules are designed to extract style and content features and subsequently reconstruct the synthesis features, as shown in Fig.~\ref{fig:main} b).
Finally, we utilize the extracted content and style features and synthesis features to establish the identity consistency module, which serves as a constraint during training to maintain identity knowledge throughout the adaptation process, as depicted in Fig.~\ref{fig:main} c).

\subsection{Identity Injection}
\label{sec3:injection}
When implementing generative models, a significant challenge is the loss of critical identity knowledge from the source domain during adaptation to the target domain.
This issue primarily arises due to the random sampling in the Mapping Network, particularly through the latent features $\mathbf{w}^i$ in the latent space $W^+$, as highlighted in prior research~\cite{chung2024style}.
To address this, we propose an identity injection module within the latent space during the adaptation process.
Our module embeds the identity knowledge from the source domain latent space vector into the target domain latent space vector, enabling the mapping network to learn and retain more source domain identity knowledge throughout training. 
This module mitigates identity loss caused by random sampling, ultimately enhancing the preservation of source domain identity in the generated target domain images.

Our module is inspired by the effectiveness of AdaIN~\cite{huang2017arbitrary}, which aligns the mean and variance of content features with those of style features, thereby facilitating the fusion of style and content information. 
In the proposed identity injection module, we align the content features of the source domain latent vector $\mathbf{w}^i_S$ with the style features of the target domain latent vector $\mathbf{w}^i_T$ to generate the identity-injected features. 
To preserve the quality of the generated images and prevent distortions from forced fusion, we further blend the identity-injected features with the original target domain latent vector. 
This refined feature then serves as the input to the target domain generator, ensuring the generated images retain high quality while incorporating the identity characteristics from the source domain. The detailed representation of our identity injection module is shown as follows:

\begin{equation}
    {\mathbf{w}^i_T}'=(1-\alpha)\cdot\mathbf{w}^i_T+\alpha\cdot\left[\sigma(\mathbf{w}^i_S)\frac{\mathbf{w}^i_T-\mu(\mathbf{w}^i_T)}{\sigma(\mathbf{w}^i_T)}+\mu(\mathbf{w}^i_S)\right]
\label{eq:injection}
\end{equation}
where $\alpha$ is a hyperparameter controlling the identity injection degree. The larger the value, the deeper the injection; the smaller the value, the shallower the injection.

Finally, through the identity injection module, we obtain target domain latent space features enriched with source domain identity knowledge ${\mathbf{w}^i_T}'$.
These features are used as input to the target domain generator to produce target domain images, which are then evaluated by the discriminator $D$. 
The mapping network $F$ is subsequently updated through backpropagation based on the final loss $\mathcal{L}_{total}$, as shown in Eq.~\ref{eq:injection-train}.
This process ensures that the mapping network $F$ retains source domain identity knowledge during parameter updates, mitigating the loss of identity information caused by random sampling.

\begin{equation}
    \Delta w_F=-lr\cdot\frac{\partial\mathcal{L}_{total}(G_T({\mathbf{w}^i_T}'),x)}{\partial w_F}\\
\label{eq:injection-train}
\end{equation}
where $\Delta w_F$ is the update parameter of the mapping network $F$, $lr$ is the learning rate of the training, and $\mathcal{L}_{total}$ refers to the total loss that combines the losses of each module and the adversarial loss $\mathcal{L}_{adv}$.

\subsection{Identity Substitution}
\label{sec3:substitution}

The pioneering work PIR~\cite{heFewshotImageGeneration2023a} introduces an innovative approach by employing an image translation module to decompose and fuse features. 
Specifically, PIR leverages the decomposed style and content features of the generated image, utilizing the translation module to cross-fuse these elements into a new image, and employs a reconstruction loss to facilitate few-shot image generation. 
However, this approach still suffers from degradation of source domain identity knowledge in the generated images. 
The core issue lies in the fact that synthesis constraints alone are insufficient to fully preserve the source domain's identity knowledge.
In response, our I\textsuperscript{2}P not only applies synthesis constraints to the corresponding features but also imposes additional constraints using the decomposed elements obtained through the decoupler, thereby enhancing the retention of source domain identity knowledge. 
In Sec.\ref{sec3:decoupler}, we introduce the style-content decoupler module for identity substitution, and in Sec.\ref{sec3:modulator}, we present the reconstruction modulator module.

\subsubsection{Style-Content Decoupler}
\label{sec3:decoupler}

In our I\textsuperscript{2}P, as shown in Fig.~\ref{fig:main} b), we utilize images generated by both the source domain generator and the target domain generator.  
We first employ the CLIP~\cite{rameshHierarchicalTextConditionalImage2022t} image encoder to extract deep features from the generated images, which serve as the original input for style and content feature extraction.
Building upon previous decomposition research~\cite{khosla2012undoing, li2017deeper, chattopadhyay2020learning, piratla2020efficient}, and given that CLIP's image encoder can extract high-quality deep features, we design a lightweight decoupler module, as shown in Fig.~\ref{fig:main} d). 
Specifically, our lightweight style-content decoupler module consists of two convolutional layers followed by a linear layer, ultimately producing a pair of linearly independent vectors representing the style and content features of the input.
Using the style-content decoupler, we extract the corresponding style and content features from images generated by the source domain, images generated by the target domain, and the training images. The detailed formulation is as follows:

\begin{equation}
\begin{aligned}
    S^i_k, C^i_k &= \mathrm{Decoupler}(\mathrm{CLIP}(x^i_k)) \\
    \text{s.t. } \forall (\alpha,\beta) &\in \mathbb{R}^2\backslash\{\mathbf{0}\},\ \alpha S^i_k + \beta C^i_k \neq \mathbf{0} \\
\end{aligned}
\end{equation}
Where $x$ represents the image, $k$ represents the source of the image, including source domain generated images, target domain generated images and raw training samples.

\subsubsection{Reconstruction Modulator}
\label{sec3:modulator}

In addition to the style-content decoupler module, we develop a reconstruction modulator module to reconstruct the decomposed style and content features into new deep features, as illustrated in Fig.~\ref{fig:main} e).
By reconstructing different deep features from various style-content pairs, our module effectively inherits and preserves the original deep feature information, allowing for the establishment of more robust constraints. 
After extracting style and content features, we leverage Adaptive Instance Normalization (AdaIN)~\cite{huang2017arbitrary} to align the mean and variance of content features with those of style features, a process that has been demonstrated as effective in prior research.
This alignment is crucial for the accurate reconstruction and fusion of style and content features. 
Consequently, AdaIN is integrated as a core component of our reconstruction modulator module, taking style and content features as inputs and generating the reconstructed features as outputs.
The characteristic reconstruction process of the modulator can be formally expressed as follows:

\begin{equation}
    M^i_{C^i_kS^i_{k'}}=\sigma(S^i_{k'})\frac{C^i_k-\mu(C^i_k)}{\sigma(C^i_k)}+\mu(S^i_{k'})
\end{equation}
Where $C^i_k$ represents the content feature, $S^i_{k'}$ represents the style feature, and $M^i_{C^i_kS^i_{k'}}$ represents the feature reconstructed by the content feature $C^i_k$ and the style feature $S^i_{k'}$. In addition, $\mu(\cdot)$ and $\sigma(\cdot)$ represent the mean and variance of the input features.

\subsection{Identity Consistency}
\label{sec3:consistency}

Based on the features obtained by the identity substitution module in the Sec.~\ref{sec3:substitution}, we model these features as probability distributions and leverage their statistical properties to construct distributional constraints for identity preservation.
To effectively preserve identity during the adaptation process, it is essential to use appropriate loss functions to constrain the distribution similarity between the style distributions, content distributions, and reconstruction distributions obtained in the identity substitution process. 
This strategy ensures that source domain-related information is retained, thereby preventing the loss of identity knowledge. 
Our module assumes that the source domain model and the target domain model should share the same identity knowledge distribution, while the target domain model and the training set should remain consistent in terms of style information. 
By maintaining these distributional consistencies, our method effectively preserves identity knowledge and enhances the quality of adaptation.

To maintain identity consistency, it is essential to focus on the content distributions $P_{C_S}$ and $P_{C_T}$, which are obtained by decomposing the source domain and target domain images, respectively.  
Similarly, to establish style consistency, the focus should be on the style distributions $P_{S_S}$ and $P_{S_R}$, derived from the target domain images and the training set. 
According to our assumption, these corresponding distributions should represent the same underlying information.
To achieve this, we employ a direct alignment strategy for these distributions. 
Specifically, we use the $\mathrm{smooth}\mbox{-}l_1$ loss to constrain these distributions, ensuring that both identity and style consistency are maintained. 
The detailed formulation is as follows:

\begin{equation}
\begin{aligned}
    \mathcal{L}_c&=\mathrm{smooth}\mbox{-}l_1(P_{C_S},P_{C_T})\\
    \mathcal{L}_s&=\mathrm{smooth}\mbox{-}l_1(P_{S_S},P_{S_R})\\
\end{aligned}
\label{eq:lcls}
\end{equation}
where $P_{C_S}$, $P_{C_T}$, $P_{S_S}$, $P_{S_R}$ represent the source domain content distribution, the target domain content distribution, the target domain style distribution, and the raw data style distribution, respectively.

Constructing content distributions $P_C$ and style distributions $P_S$ constraints alone is insufficient to fully preserve the source domain identity knowledge and facilitate effective style learning. 
In our module, content distributions $P_C$ and style distributions $P_S$ constraints are implemented by directly establishing correspondences between distributions and aligning content and style elements. 
However, solely relying on these constraints can result in overly restrictive relationships between distributions, leading to distortions and non-smooth transitions in the final target domain generator, which ultimately degrades the quality of the generated images.

To address these issues, we introduce a synthesis consistency constraint to mitigate the limitations of direct content and style distribution constraints while ensuring that the synthesis distributions $P_M$ retain the integrity of the deep feature distributions obtained through CLIP. 
Based on our analysis, the synthesis distributions $P_M$ of interest are those in the reconstructed image distributions that are similar to the generated image distributions in the target domain in terms of both style and content—specifically, $P_{M_{C_SS_R}}$, $P_{M_{C_TS_R}}$, and $P_{M_{C_SS_T}}$.
These distributions are reconstructed by the reconstruction modulator and represent the fused deep feature distributions, effectively maintaining the balance between identity preservation and style adaptation.

Since the content and style losses are directly established between these content and style distributions, it is important to consider the nonlinear relationships between the reconstructed synthesis distributions when constructing the synthesis consistency constraint.
Unlike the content and style constraints, which focus on direct distribution alignment, the synthesis constraint aims to ensure that the reconstructed distributions align with the target distributions in terms of spatial directionality (i.e., the directionality of identity) rather than solely focusing on numerical alignment.
To achieve this, we employ cosine similarity to constrain the three synthesis distributions, ensuring that they remain as similar as possible throughout the training process. 
The proposed synthesis constraint is defined by calculating pairwise cosine similarities between the three distributions, specifically expressed as:

\begin{equation}
\begin{aligned}
    \mathcal{L}_r=\sum_{l,l'}\mathrm{sim}(P_{M_{l}},P_{M_{l'}})
\end{aligned}
\label{eq:lr}
\end{equation}
where $P_M$ represent the synthesis distributions, $\mathrm{sim}(\cdot,\cdot)$ represent the cosine similarity, $l\neq l'$ and $l,l'\in\{C_SS_R, C_TS_R, C_SS_T\}$ 

Finally, the loss function of our optimization objective is expressed as follows:

\begin{equation}
    \mathcal{L}_{total}=\mathcal{L}_{adv}+\lambda\cdot(\mathcal{L}_{c}+\mathcal{L}_{s}+\mathcal{L}_{r})
\label{eq:total-loss}
\end{equation}
where $\lambda$ is a hyperparameter.

\begin{figure*}[htbp]
    \centering
    \includegraphics[width=1\textwidth]{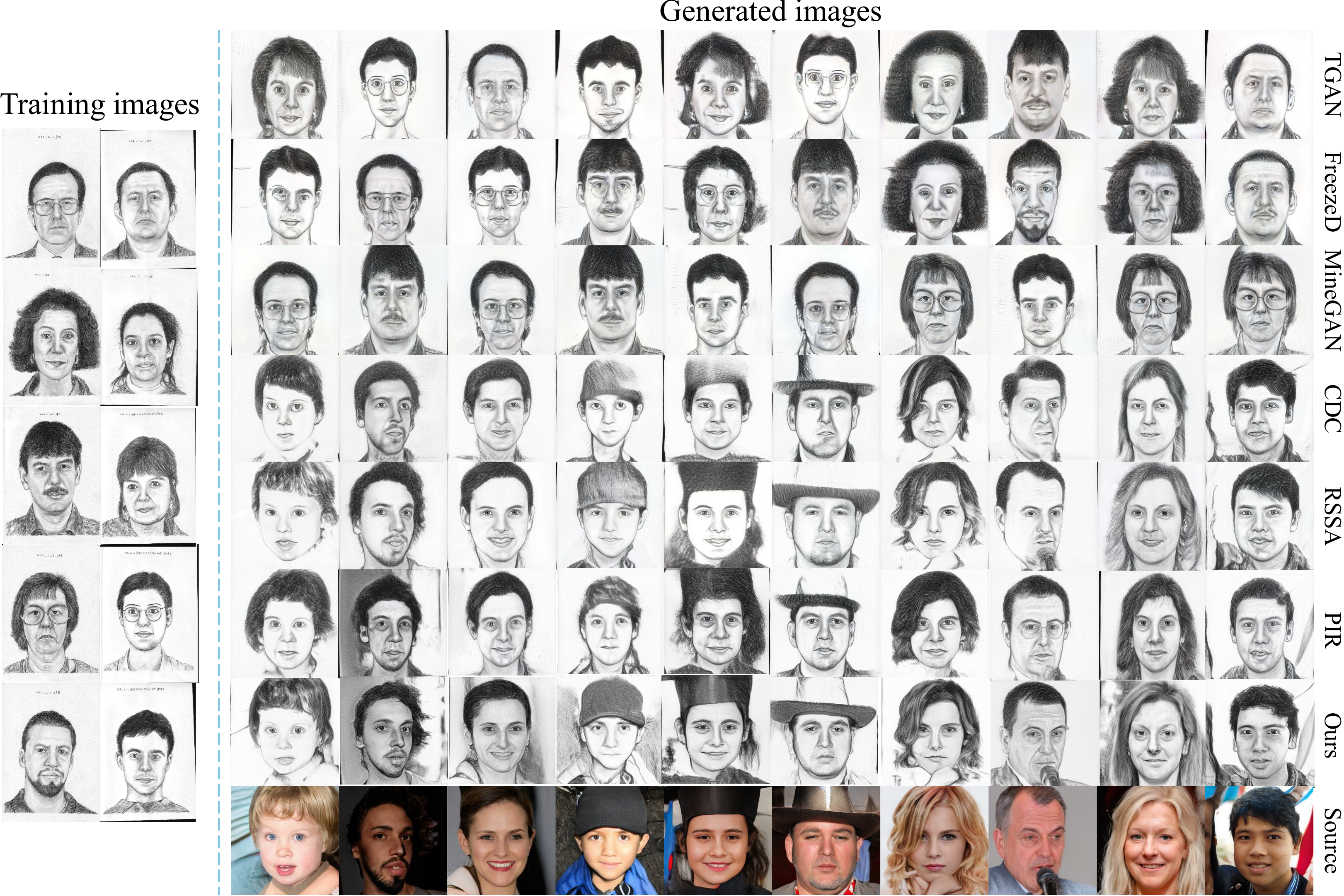}
    \caption{
        Cross-domain adaptation performance comparison on Flickr-Faces-HQ (FFHQ)$\rightarrow$Sketches task. 
        (a) Baseline methods (TGAN, FreezeD, MineGAN) exhibit overfitting artifacts in 10-shot setting; 
        (b) Recent approaches (CDC, RSSA, PIR) show improved alignment but suffer from content distortion or style inconsistency; 
        (c) Our I\textsuperscript{2}P framework maintains structural fidelity and stylistic coherence through identity-preserved adaptation. 
        Visualizations demonstrate our method's capability to generate high-fidelity sketches while preserving source domain identity features from identical latent codes, significantly outperforming state-of-the-art methods in both quality and cross-domain consistency.
    }
    \label{fig:exp-main}
\end{figure*}

\section{Experiments}
\label{sec4:experiments}

In this section, we demonstrate the effectiveness of I\textsuperscript{2}P in few-shot settings through both qualitative and quantitative comparisons with several baseline methods, including TGAN~\cite{wang2018transferring}, FreezeD~\cite{mo2020freeze}, MineGAN~\cite{wang2020minegan}, CDC~\cite{ojhaFewshotImageGeneration2021c}, RSSA~\cite{xiao2022few}, PIR~\cite{heFewshotImageGeneration2023a}, and SGP~\cite{pan2024few}. It is important to note that SGP~\cite{pan2024few} has not open-sourced its code, which limits our ability to conduct experiments for comparison.

We adapt StyleGANv2~\cite{karras2020analyzing}, pre-trained on high-quality, large datasets such as Flickr-Faces-HQ (FFHQ)~\cite{karrasStyleBasedGeneratorArchitecture2019g}, LSUN-Churches~\cite{yu2015lsun}, LSUN-Cars~\cite{yu2015lsun}, AFHQ-Cat~\cite{choi2020starganv2} and AFHQ-Dog~\cite{choi2020starganv2}. The source generative models are then adapted to various target domains, including Sketches~\cite{wang2008face}, Metfaces~\cite{karras2020training}, FFHQ-Babies~\cite{karrasStyleBasedGeneratorArchitecture2019g}, FFHQ-Sunglasses~\cite{karrasStyleBasedGeneratorArchitecture2019g}, village paintings by VanGogh~\cite{ojhaFewshotImageGeneration2021c}, Haunted-houses~\cite{ojhaFewshotImageGeneration2021c}, Haunted-cars~\cite{ojhaFewshotImageGeneration2021c}, landscape-cars, etc. Additionally, we adapt the pre-trained StyleGANv2~\cite{karras2020analyzing} generator to target domains with a batch size of 4 on a single NVIDIA RTX 3090 with 5002 iterations.

\subsection{Performance evaluation}
\label{sec4:evaluation}

\subsubsection{Qualitative comparison}
\label{sec4:qualitative}
\begin{figure*}[htbp]
    \centering
    \includegraphics[width=1\textwidth]{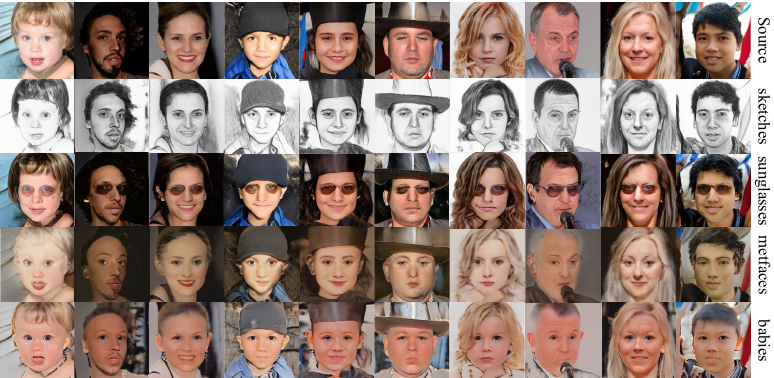}
    \caption{
        The cross-domain adaptation results of our I\textsuperscript{2}P are shown in the conversion scenarios between FFHQ and four different target domains: FFHQ$\rightarrow$Sketches, FFHQ$\rightarrow$Sunglasses, FFHQ$\rightarrow$Metfaces and FFHQ$\rightarrow$Babies,
        Qualitative results for each target domain illustrate the ability of our I\textsuperscript{2}P to adapt the same source domain model to different target domains while preserving domain-invariant features.
        }
    \label{fig:exp-ffhq}
\end{figure*}

\begin{figure*}[htbp]
    \centering
    \includegraphics[width=1\textwidth]{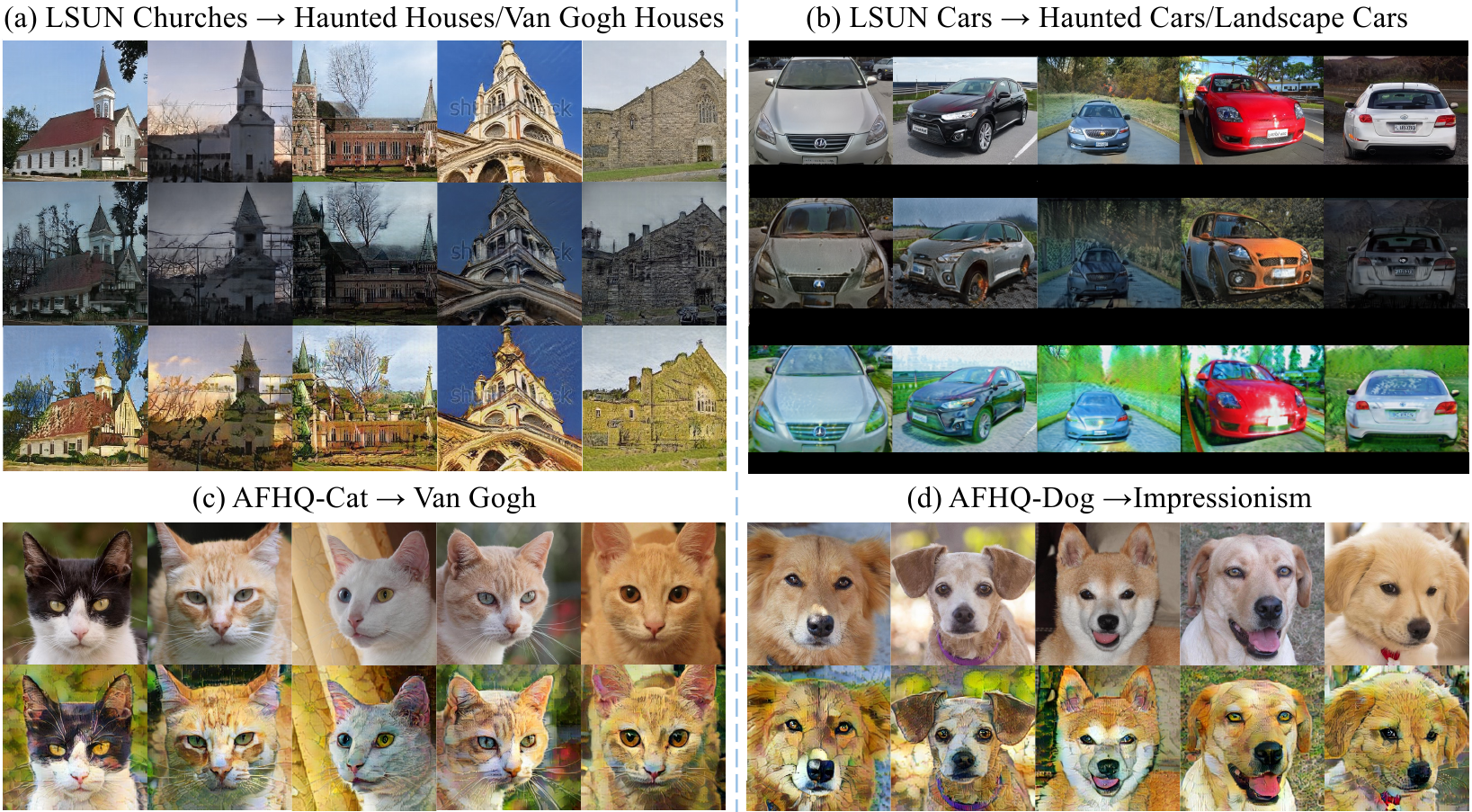}
    \caption{
        The cross-domain adaptation results of our I\textsuperscript{2}P are shown in four different conversion scenarios:
        (a) LSUN-Churches $\rightarrow$ Haunted-Houses/VanGogh(top left);
        (b) LSUN-Cars $\rightarrow$ Haunted-Cars/Landscape-Cars(top right);
        (c) AFHQ-Cat $\rightarrow$ VanGogh(bottom left);
        (d) AFHQ-Dog $\rightarrow$ Impressionism(bottom right).
        Each quadrant illustrates our I\textsuperscript{2}P's capability in preserving domain-invariant features while adapting to diverse artistic styles.
    }
    \label{fig:exp-other}
\end{figure*}

\begin{figure}[htbp]
    \centering
    \includegraphics[width=1\columnwidth]{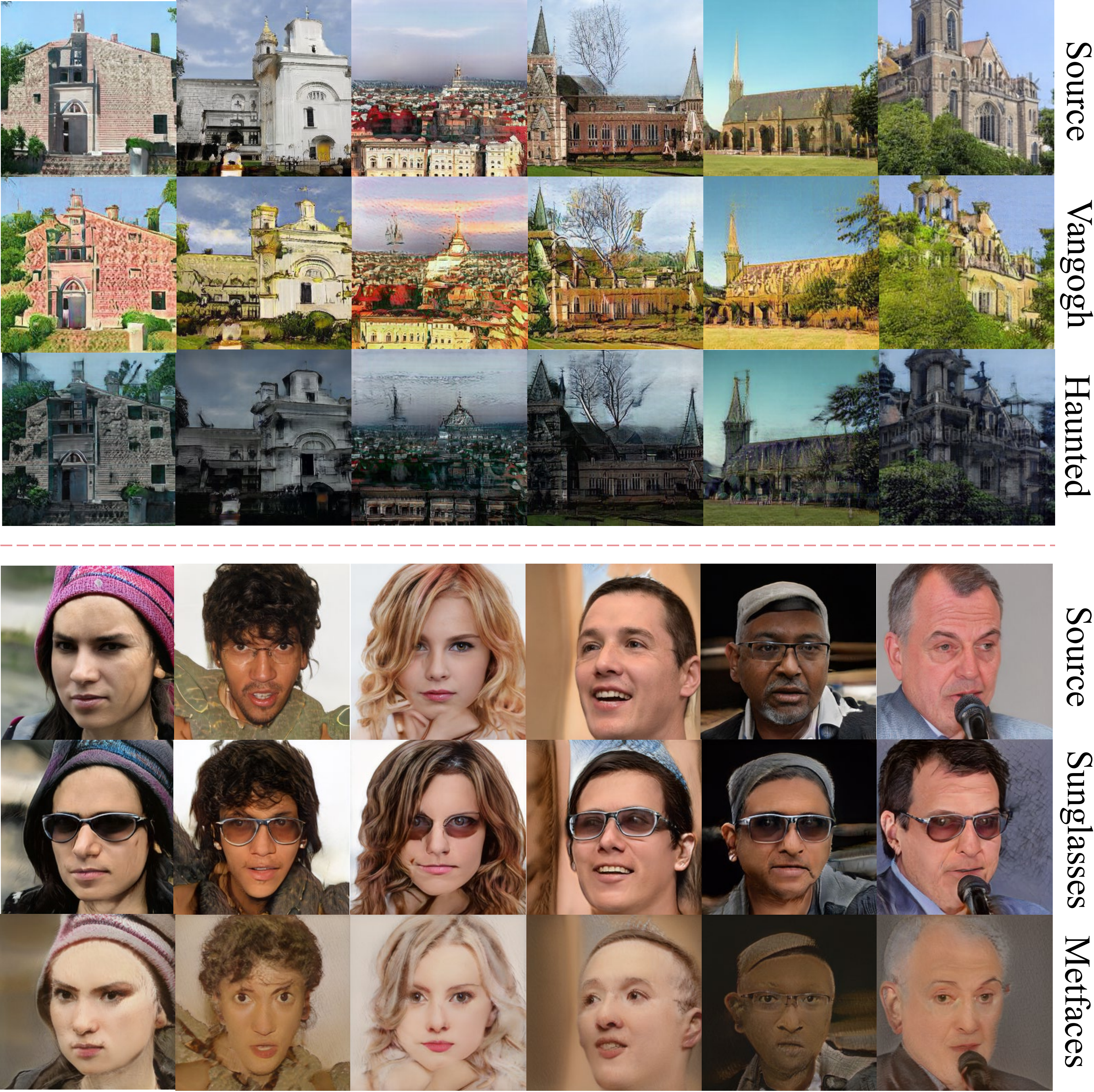} 
    \vspace{-5mm}
    \caption{The \textbf{5-shot} generation results validate the capability of our I\textsuperscript{2}P in cross-domain adaptation on fewer samples, achieving faithful style transfer while maintaining identity knowledge with the source domain.}
    \label{fig:5-shot}
\end{figure}

As shown in Fig.~\ref{fig:exp-main}, this qualitative comparison highlights the performance of our method, I\textsuperscript{2}P, against a range of mainstream approaches for the Flickr-Faces-HQ (FFHQ) to Sketches task. Earlier methods such as TGAN, FreezeD, and MineGAN exhibit varying degrees of overfitting, especially when the available training data is limited to only 10-shot.
Although these techniques perform reasonably well with larger few-shot datasets (e.g., over 100-shot), their effectiveness drops significantly as the data becomes sparser. In contrast, more recent sota methods, such as CDC, RSSA, and PIR, demonstrate improve resilience in extremely few-shot settings, with a clearer visual correspondence between source and target domain images.
However, these methods still have notable drawbacks. CDC and PIR suffer from issues like overfitting and content distortion. While RSSA retains more identity information, it results in a lack of stylization in the generated images.
In contrast, our method, I\textsuperscript{2}P, consistently outperforms these approaches by effectively preserving the source domain identity knowledge, even in 10-shot (Fig.~\ref{fig:exp-main}) and 5-shot (Fig.~\ref{fig:5-shot}) scenarios across various source and target domains, with no significant signs of identity degradation or overfitting.
In addition, as shown in the Fig.~\ref{fig:exp-ffhq} and Fig.~\ref{fig:exp-other}, I\textsuperscript{2}P also conducts qualitative experiments on several mainstream source and target domains (FFHQ$\rightarrow$FFHQ-babies, FFHQ$\rightarrow$MetFaces, FFHQ$\rightarrow$Sketches, FFHQ$\rightarrow$FFHQ-sunglasses, LSUN-Churches$\rightarrow$Haunted-Houses, LSUN-Churches $\rightarrow$VanGogh, LSUN-Cars$\rightarrow$Haunted-Cars, LSUN-Cars$\rightarrow$Landscape-Cars, AFHQ-Cat$\rightarrow$VanGogh, AFHQ-Dog$\rightarrow$Impressionism). The results show that regardless of the different source domain generation models or the changes in the target domain training data, our method can always excellently preserve the identity information of the source domain while effectively transferring the style elements of the target domain, even if the data is very limited. This ability to achieve high-quality visual transfer with minimal target domain data highlights the robustness and versatility of our method.

\subsubsection{Quantitative comparison}
\label{sec4:quantitative}

To comprehensively evaluate the quality and diversity of the synthesized images, we conduct experiments using the FID score~\cite{heusel2017gans} between the generated target domain images with the raw target domain images. 
Meanwhile, to assess the perceptual diversity of images within the same domain, we also utilize the intra-cluster pairwise LPIPS distance (Intra-LPIPS)~\cite{zhang2018unreasonable, ojhaFewshotImageGeneration2021c} between generated source domain images with generated target domain images.
The specific calculation formulas of FID score and Intra-LPIPS are as follows:

\begin{equation}
\begin{aligned}
    \mathbf{FID}^2 = \Vert\mu_1-\mu_2\Vert^2_2+ \mathbf{Tr}(\Sigma_1+ \Sigma_2-2\cdot(\Sigma_1\Sigma_2)^{\frac{1}{2}})
\end{aligned}
\label{eq:fid}
\end{equation}
Where $\mu_i$ represents the mean vector of the real data distribution and the generative model, $\Sigma_i$ represents the covariance matrix of the real data and the generative model.

\begin{equation}
\begin{aligned}
    \mathbf{LPIPS}(x,x_0)=\sum_l \dfrac{1}{H_l W_l} \sum_{h,w} || w_l \odot ( \hat{y}_{hw}^l - \hat{y}_{0hw}^l ) ||_2^2
\end{aligned}
\label{eq:lpips}
\end{equation}
Where $x$ and $x_0$ represent two images. Intra-LPIPS is calculated by dividing the image set into multiple clusters and averaging the pairwise LPIPS~\cite{zhang2018unreasonable} (Eq.~\ref{eq:lpips}) within the same cluster.

We select four datasets, FFHQ-Babies~\cite{karrasStyleBasedGeneratorArchitecture2019g}, MetFaces~\cite{karras2020training}, Sketches~\cite{wang2008face}, and FFHQ-Sunglasses~\cite{karrasStyleBasedGeneratorArchitecture2019g}, that provide sufficient data for comprehensive evaluation.
For each dataset, models trained with different methods generate a number of samples equivalent to the original dataset size, which are then used to calculate the FID score for each method. 
As illustrated in Table~\ref{table:experiment}, our I\textsuperscript{2}P consistently achieved the lowest FID score across all four datasets, indicating its superior ability to replicate the true distribution of the source domain.
However, since FID alone does not account for model collapse, we also assessed the the perceptual diversity of images using the Intra-LPIPS distance.
Due to poor identity preservation observed in qualitative experiments, TGAN, FreezeD, and MineGAN are excluded from the Intra-LPIPS comparison.
As shown in Table~\ref{table:experiment}, our method achieve a higher average Intra-LPIPS distance compared to other approaches, reinforcing the significant advantage of I\textsuperscript{2}P in  maintaining image diversity. 

For assessing identity preservation as discussed in our paper, we incorporat three advanced indicators from image generation research to comprehensively assess image generation quality: DINO~\cite{oquab2023dinov2}, CLIP-I~\cite{rameshHierarchicalTextConditionalImage2022t}, and CLIP-T~\cite{rameshHierarchicalTextConditionalImage2022t}. 
The first two metrics perform cross-domain comparison between source and target images through deep feature analysis. 
Specifically, DINO v2 and CLIP's image encoder are employed to extract high-dimensional feature representations from both domains, followed by cosine similarity measurement in their respective latent spaces. 
While both metrics evaluate feature-level correspondence, they emphasize different aspects: DINO primarily captures structural and compositional similarities through its self-supervised vision transformer, whereas CLIP-I focuses on semantic alignment leveraging its multimodal training objective. 
We also introduce CLIP-T as a novel stylization metric to quantitatively evaluate style transfer effectiveness. 
This measurement employs a text-guided evaluation paradigm using the standardized prompt template ``a \textbf{\{source domain name\}} image in \textbf{\{target domain name\}}''. 
The CLIP text encoder processes this prompt to generate text embeddings, which are then compared with image embeddings of target domain outputs through cosine similarity. 

As shown in Table~\ref{tab:add-main}, the results of DINO, CLIP-I and CLIP-T quantitative experiments of our method and three baseline methods, adapted from the source domain FFHQ to four target domains FFHQ-Babies, MetFaces, Sketches, and FFHQ-Sunglasses. 
It can be observed that our method has taken the lead in most of the sota adaptation results. 
The results of DINO, CLIP-I and CLIP-T further show that our I\textsuperscript{2}P better preserves the identity knowledge of the source domain during the adaptation process and also achieves stylized unification.
These results underscore the robustness and versatility of our method across different datasets and target domains.

\begin{table*}[htbp]
    \normalsize
    \centering
    \caption{
    Quantitative evaluation of methods by FID score($\downarrow$) and Intra-LPIPS distance($\uparrow$).The source domain comes from FFHQ, and the target domain dataset comes from FFHQ-Babies, Metafaces, sketches, and FFHQ-Sunglasses. 
    }
    \begin{tabular}{c|cccc|cccc}
    \hline
                            & \multicolumn{4}{c|}{FID$\downarrow$}                                                                                                                                          & \multicolumn{4}{c}{Intra-LPIPS$\uparrow$}                                                                                                                                         \\
    \multirow{-2}{*}{Method} & babies                                 & metfaces                               & sketches                               & sunglasses                             & babies                                 & metfaces                               & sketches                               & sunglasses                             \\ \hline
    TGAN(2018)                     & \cellcolor[HTML]{D1E3EF}104.79         & \cellcolor[HTML]{B6D5E4}76.81          & \cellcolor[HTML]{B4D4E3}53.42          & \cellcolor[HTML]{BAD7E6}55.61          & $-$                                    & $-$                                    & $-$                                    & $-$                                    \\
    FreezeD(2020)                  & \cellcolor[HTML]{E3EDF7}113.01         & \cellcolor[HTML]{C5DDEA}79.33          & \cellcolor[HTML]{ACCFE0}50.53          & \cellcolor[HTML]{ADD0E0}51.29          & $-$                                    & $-$                                    & $-$                                    & $-$                                    \\
    MineGAN(2020)                  & \cellcolor[HTML]{C8DEEB}100.15         & \cellcolor[HTML]{E3EDF7}84.27          & \cellcolor[HTML]{E3EDF7}69.77          & \cellcolor[HTML]{E3EDF7}68.91          & $-$                                    & $-$                                    & $-$                                    & $-$                                    \\
    CDC(2021)                      & \cellcolor[HTML]{92C1D5}74.39          & \cellcolor[HTML]{C9DFEC}80.00          & \cellcolor[HTML]{9FC8DA}45.67          & \cellcolor[HTML]{92C1D5}42.13          & \cellcolor[HTML]{FFE8FF}0.450          & \cellcolor[HTML]{FAE1FC}0.440          & \cellcolor[HTML]{DEBCE7}0.605          & \cellcolor[HTML]{FFE8FF}0.312          \\
    RSSA(2022)                     & \cellcolor[HTML]{95C2D6}75.67          & \cellcolor[HTML]{B8D6E5}77.06          & \cellcolor[HTML]{B8D5E5}54.58          & \cellcolor[HTML]{98C5D8}44.35          & \cellcolor[HTML]{DBB8E5}0.605          & \cellcolor[HTML]{E2C2EB}0.488          & \cellcolor[HTML]{D8B4E3}0.635          & \cellcolor[HTML]{DBB8E5}0.400          \\
    PIR(2024)                      & \cellcolor[HTML]{8ABDD2}70.50          & \cellcolor[HTML]{ACCFE0}75.02          & \cellcolor[HTML]{9DC7DA}45.01          & \cellcolor[HTML]{98C4D7}44.05          & \cellcolor[HTML]{E2C1EA}0.576          & \cellcolor[HTML]{FFE8FF}0.429          & \cellcolor[HTML]{FFE8FF}0.436          & \cellcolor[HTML]{E1C0E9}0.385          \\
    SGP(2024)                      & \cellcolor[HTML]{8ABDD2}70.76          & $-$                                    & \cellcolor[HTML]{BBD7E6}55.82          & \cellcolor[HTML]{91C0D4}41.76          & \cellcolor[HTML]{D8B4E3}0.618          & $-$                                    & \cellcolor[HTML]{E7C9EE}0.604          & $-$                                    \\ \hline
    I\textsuperscript{2}P(Ours)                & \cellcolor[HTML]{8ABDD2}\textbf{70.39} & \cellcolor[HTML]{8ABDD2}\textbf{69.20} & \cellcolor[HTML]{8ABDD2}\textbf{38.16} & \cellcolor[HTML]{8ABDD2}\textbf{39.40} & \cellcolor[HTML]{D5B1E1}\textbf{0.627} & \cellcolor[HTML]{D5B1E1}\textbf{0.513} & \cellcolor[HTML]{D5B1E1}\textbf{0.646} & \cellcolor[HTML]{D5B1E1}\textbf{0.412} \\ \hline
    \end{tabular}
    \label{table:experiment}
\end{table*}

\begin{table*}[htbp]
    \normalsize
    \centering
    \caption{Quantitative evaluation of methods by DINO score($\uparrow$), CLIP-I score($\uparrow$) and CLIP-T score($\uparrow$).The source domain comes from FFHQ, and the target domain dataset comes from FFHQ-Babies, Metafaces, sketches, and FFHQ-Sunglasses.}
    \begin{tabular}{c|cccc|cccc|cccc}
    \hline
                            & \multicolumn{4}{c|}{DINO$\uparrow$}                                                                                                                                               & \multicolumn{4}{c|}{CLIP-I$\uparrow$}                                                                                                                                             & \multicolumn{4}{c}{CLIP-T$\uparrow$}                                                                                                                                        \\
    \multirow{-2}{*}{Method} & b                                      & m                                      & sk                                     & su                                           & b                                      & m                                      & sk                                     & su                                           & b                                      & m                                      & sk                                     & su                                     \\ \hline
    CDC(2021)                      & \cellcolor[HTML]{FFFFFF}0.620          & \cellcolor[HTML]{D8E9F0}0.601          & \cellcolor[HTML]{FFFFFF}0.560          & \cellcolor[HTML]{FFFFFF}0.744                & \cellcolor[HTML]{DCBDE6}0.601          & \cellcolor[HTML]{FFFFFF}0.548          & \cellcolor[HTML]{FFFFFF}0.556          & \cellcolor[HTML]{FFFFFF}0.640                & \cellcolor[HTML]{FFEEE1}{\ul 0.166}    & \cellcolor[HTML]{FFFBF7}0.197          & \cellcolor[HTML]{FFFFFF}0.203          & \cellcolor[HTML]{FFFFFF}0.186          \\
    RSSA(2022)                     & \cellcolor[HTML]{BFDBE7}{\ul 0.630}    & \cellcolor[HTML]{D3E6EE}{\ul 0.602}    & \cellcolor[HTML]{C6DFE9}{\ul 0.580}    & \cellcolor[HTML]{8ABDD2}\textbf{0.788}       & \cellcolor[HTML]{D8B6E3}{\ul 0.602}    & \cellcolor[HTML]{EBDAF1}0.555          & \cellcolor[HTML]{E1C7EA}{\ul 0.601}    & \cellcolor[HTML]{D5B1E1}\textbf{0.675}       & \cellcolor[HTML]{FFF1E7}0.166          & \cellcolor[HTML]{FFFFFF}0.195          & \cellcolor[HTML]{FFF3EA}0.205          & \cellcolor[HTML]{FFECDE}\textbf{0.192}    \\
    PIR(2024)                      & \cellcolor[HTML]{FAFCFD}0.621          & \cellcolor[HTML]{FFFFFF}0.592          & \cellcolor[HTML]{E6F1F6}0.568          & \cellcolor[HTML]{EBF4F7}0.752                & \cellcolor[HTML]{FFFFFF}0.588          & \cellcolor[HTML]{E7D1EE}{\ul 0.557}    & \cellcolor[HTML]{F8F2FA}0.567          & \cellcolor[HTML]{FDFBFE}0.642                & \cellcolor[HTML]{FFFFFF}0.162          & \cellcolor[HTML]{FFF5EE}{\ul 0.199}    & \cellcolor[HTML]{FFEFE4}{\ul 0.205}    & \cellcolor[HTML]{FFF7F0}0.189          \\
    I\textsuperscript{2}P(Ours)                      & \cellcolor[HTML]{8ABDD2}\textbf{0.639} & \cellcolor[HTML]{8ABDD2}\textbf{0.618} & \cellcolor[HTML]{8ABDD2}\textbf{0.600} & \cellcolor[HTML]{B8D7E4}{\ul 0.771} & \cellcolor[HTML]{D5B1E1}\textbf{0.603} & \cellcolor[HTML]{D5B1E1}\textbf{0.563} & \cellcolor[HTML]{D5B1E1}\textbf{0.618} & \cellcolor[HTML]{DFC2E8}{\ul 0.667} & \cellcolor[HTML]{FFECDE}\textbf{0.167} & \cellcolor[HTML]{FFECDE}\textbf{0.202} & \cellcolor[HTML]{FFECDE}\textbf{0.206} & \cellcolor[HTML]{FFEEE1}{\ul 0.192} \\ \hline
    \end{tabular}
    \label{tab:add-main}
\end{table*}

\subsection{Ablation study}
\label{sec4:ablation}

\subsubsection{Effect of the Identity Injection and Preservation}
\label{sec4:effect-module}

\begin{figure}[htbp]
    \centering
    \includegraphics[width=1\columnwidth]{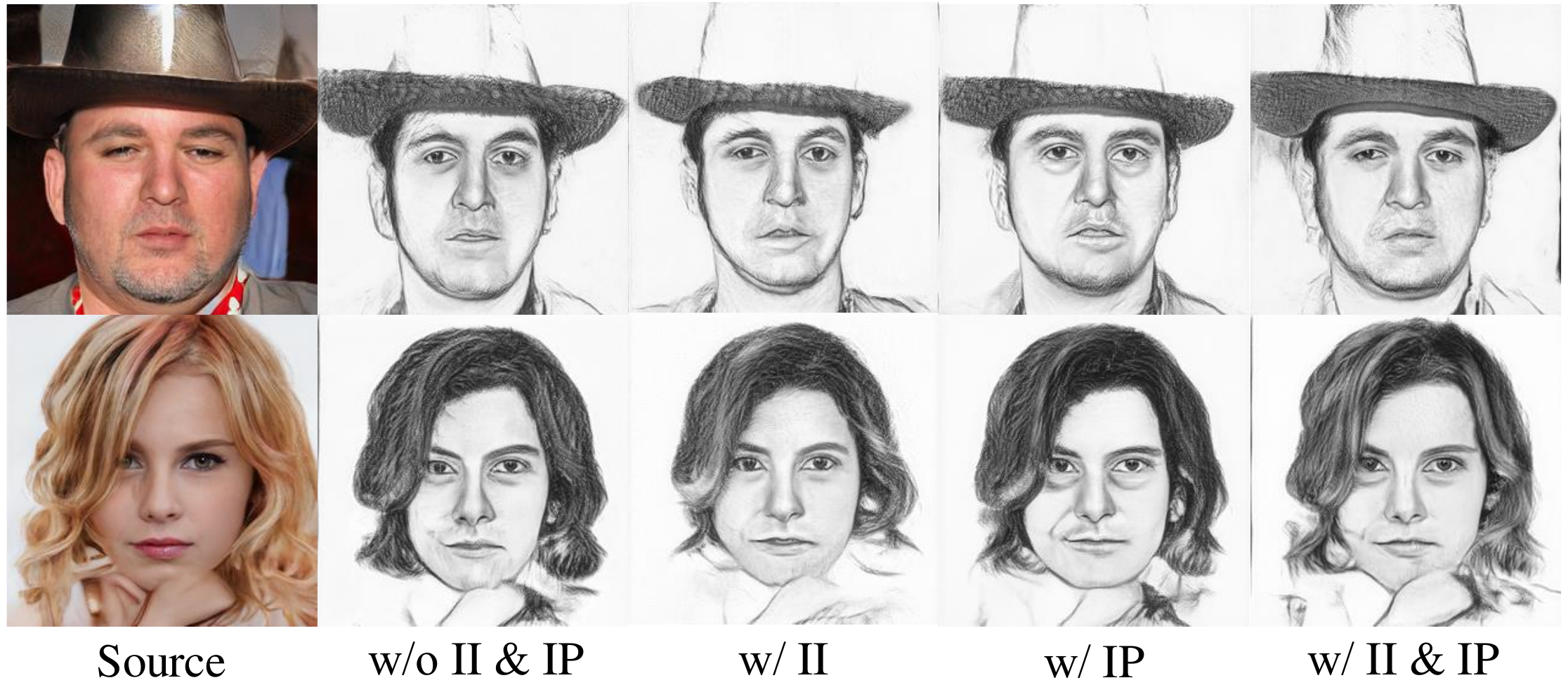} 
    \vspace{-5mm}
    \caption{Qualitative ablation of the I\textsuperscript{2}P's modules. II refers to the identity injection module. IP refers to the identity preservation, including identity substitution and identity consistency.}
    \label{fig:ablation}
\end{figure}

We perform ablation experiments on the identity injection module to evaluate its effectiveness. 
As shown in Fig.~\ref{fig:ablation}, focusing on areas such as the chin, mouth, and other facial features, the module successfully retains a significant amount of source domain identity knowledge in the generated images. 
This is further supported by the quantitative results in Table~\ref{tab:sup-ablation1-fid} and Table~\ref{tab:sup-ablation1-lpips}, which demonstrate the robustness of the identity preservation module in replicating the true distribution of the source domain.

Simultaneously, we conduct ablation experiments on the identity preservation module to verify its effectiveness. 
This module comprises the identity substitution and identity consistency components. 
As shown in Fig.~\ref{fig:ablation}, the identity preservation module significantly reduces identity knowledge loss and enhances the detailed features of the generated images in the target domain, particularly in areas such as the eye bags, nasolabial folds, and beard. 
These fundamental improvements originate from our novel identity substitution mechanism and rigorous consistency constraints, which proactively maintain domain-agnostic characteristics throughout the adaptation process.
The quantitative data in Table~\ref{tab:sup-ablation1-fid} and Table~\ref{tab:sup-ablation1-lpips} further supports this conclusion, highlighting the critical role of the identity preservation module in enhancing both qualitative and quantitative outcomes.

\begin{table}[htbp]
    \normalsize
    \centering
    \caption{Quantitative ablation experiments on FID$\downarrow$ of our I\textsuperscript{2}P method. II refers to the identity injection module. IP refers to the identity preservation, including identity substitution and identity consistency.}
    \label{tab:sup-ablation1-fid}
    \begin{tabular}{ccccccc}
    \hline
                        & \multicolumn{2}{c}{IP} & \multicolumn{4}{c}{FID$\downarrow$}                                                                                                                                           \\ \cline{4-7} 
    \multirow{-2}{*}{II} & $L_r$    & $L_c$\&$L_s$    & b                                 & m                               & sk                               & su                             \\ \hline
    \ding{55}                    & \ding{55}       & \ding{55}            & \cellcolor[HTML]{FFFFFF}79.99          & \cellcolor[HTML]{FFFFFF}77.15          & \cellcolor[HTML]{FFFFFF}44.55          & \cellcolor[HTML]{FFFFFF}45.05          \\
    \ding{55}                    & \ding{51}       & \ding{51}            & \cellcolor[HTML]{E1EEF3}77.54          & \cellcolor[HTML]{E4F0F4}75.33          & \cellcolor[HTML]{9FC9DA}39.36          & \cellcolor[HTML]{E7F1F5}43.55          \\
    \ding{51}                    & \ding{55}       & \ding{55}            & \cellcolor[HTML]{BDDAE5}74.63          & \cellcolor[HTML]{D2E5ED}74.05          & \cellcolor[HTML]{9BC6D8}39.09          & \cellcolor[HTML]{A4CBDC}39.40          \\
    \ding{51}                    & \ding{55}       & \ding{51}            & \cellcolor[HTML]{B9D7E4}74.27          & \cellcolor[HTML]{C4DDE8}73.09          & \cellcolor[HTML]{90C0D4}38.52          & \cellcolor[HTML]{94C3D6}38.43          \\
    \ding{51}                    & \ding{51}       & \ding{55}            & \cellcolor[HTML]{A5CCDC}72.67          & \cellcolor[HTML]{8ABDD2}69.35          & \cellcolor[HTML]{8ABDD2}38.20          & \cellcolor[HTML]{91C1D4}38.21          \\
    \ding{51}                    & \ding{51}       & \ding{51}            & \cellcolor[HTML]{8ABDD2}\textbf{70.39} & \cellcolor[HTML]{8CBED2}\textbf{69.20} & \cellcolor[HTML]{8ABDD2}\textbf{38.16} & \cellcolor[HTML]{8ABDD2}\textbf{37.75} \\ \hline
    \end{tabular}
\end{table}

\subsubsection{Effect of the Identity Consistency}
\label{sec4:effect-optimization}

\begin{figure}[htbp]
    \centering
    \includegraphics[width=1\columnwidth]{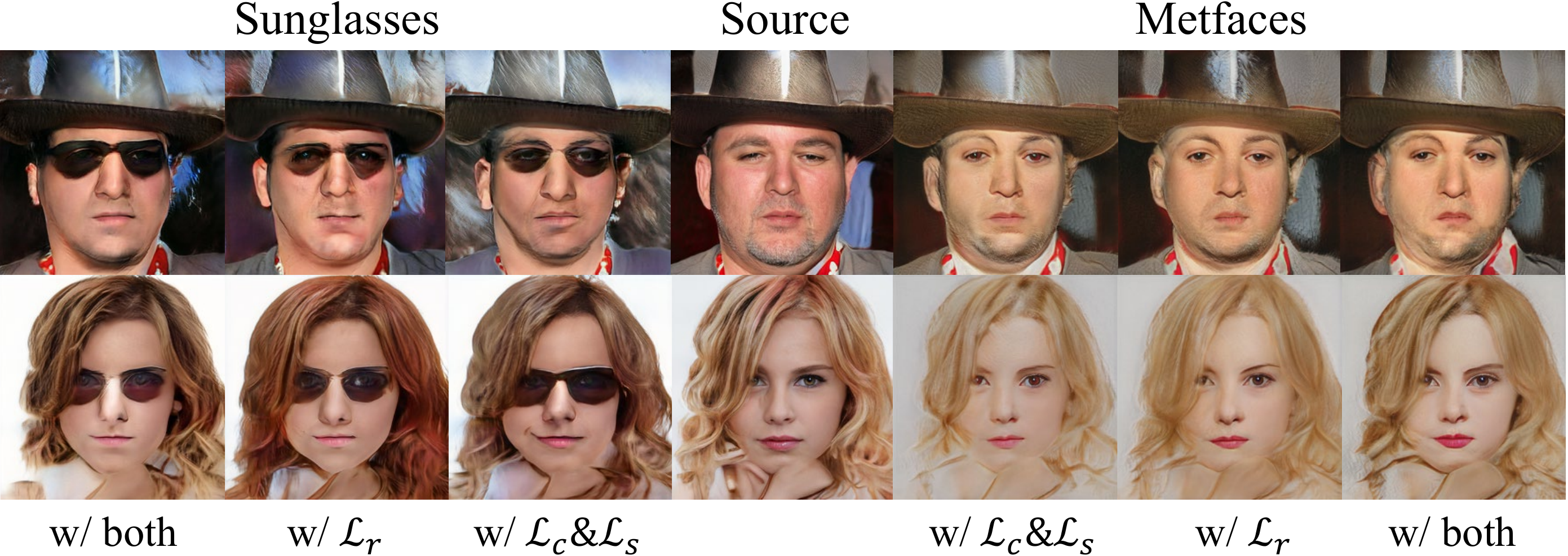}
    \vspace{-5mm}
    \caption{Qualitative ablation experiments of our I\textsuperscript{2}P on two types of losses in the Identity Preservation module: synthesis constraint and content constraint \& style constraint}
    \label{fig:sup-ablation1}
\end{figure}

We conducte ablation experiments to analyze the internal structure of the identity consistency module, specifically focusing on the relationship between $\mathcal{L}_r$ (Eq.~\ref{eq:lr}) and $\mathcal{L}_c \& \mathcal{L}_s$ (Eq.~\ref{eq:lcls}). 
The systematic ablation studies in Fig.~\ref{fig:sup-ablation1} and Tables~\ref{tab:sup-ablation1-fid}-\ref{tab:sup-ablation1-lpips} reveal the synergistic relationship between synthesis constraint ($\mathcal{L}_r$) and content-style constraints ($\mathcal{L}_c$\&$\mathcal{L}_s$), where their coordinated optimization enables simultaneous achievement of identity consistency and style fidelity.
In the qualitative analysis, depicted in Fig.~\ref{fig:sup-ablation1}, it is evident that using only $\mathcal{L}_c \& \mathcal{L}_s$ results in noticeable distortions and oversharpening in facial contours, as well as exaggerated lines and shadows around the eye sockets. This finding supports our hypothesis that overly strong constraints can introduce undesirable distortion in the generated images. 

Conversely, when only $\mathcal{L}_r$ is enforced, the generated images exhibit a slight loss of identity information, manifesting as subtle artifacts such as shadows. Nevertheless, the overall quality remains comparable to the results obtained when both $\mathcal{L}_r$ and $\mathcal{L}_c \& \mathcal{L}_s$ are enforced simultaneously, which aligns with the quantitative outcomes.

When these two losses are combined at an appropriate ratio, the generated images in the target domain retain a high degree of source domain identity knowledge without introducing distortions caused by overly strong constraints. This balance further demonstrates the effectiveness of our method in preserving identity while maintaining high visual quality.

\begin{table}[htbp]
    \normalsize
    \centering
    \caption{Quantitative experiments on Intra-LPIPS$\uparrow$ of our I\textsuperscript{2}P method. II refers to the identity injection module. IP refers to the identity preservation, including identity substitution and identity consistency.}
    \label{tab:sup-ablation1-lpips}
    \begin{tabular}{ccccccc}
    \hline
                        & \multicolumn{2}{c}{IP} & \multicolumn{4}{c}{Intra-LPIPS$\uparrow$}                                                                                                                                   \\ \cline{4-7} 
    \multirow{-2}{*}{II} & $L_r$    & $L_c\&L_s$   & b                                      & m                                      & sk                                     & su                                     \\ \hline
    \ding{55}                    & \ding{55}       & \ding{55}            & \cellcolor[HTML]{FFFFFF}0.449          & \cellcolor[HTML]{FFFFFF}0.396          & \cellcolor[HTML]{FFFFFF}0.486          & \cellcolor[HTML]{FFFFFF}0.349          \\
    \ding{55}                    & \ding{51}       & \ding{51}            & \cellcolor[HTML]{E7D3EE}0.551          & \cellcolor[HTML]{E1C6E9}0.482          & \cellcolor[HTML]{E4CCEC}0.592          & \cellcolor[HTML]{F5EDF8}0.364          \\
    \ding{51}                    & \ding{55}       & \ding{55}            & \cellcolor[HTML]{EAD7F0}0.542          & \cellcolor[HTML]{E7D3EE}0.463          & \cellcolor[HTML]{E7D2EE}0.579          & \cellcolor[HTML]{F6EEF9}0.363          \\
    \ding{51}                    & \ding{55}       & \ding{51}            & \cellcolor[HTML]{DCBDE6}0.601          & \cellcolor[HTML]{DDBFE7}0.493          & \cellcolor[HTML]{DCBDE6}0.622          & \cellcolor[HTML]{E9D7F0}0.382          \\
    \ding{51}                    & \ding{51}       & \ding{55}            & \cellcolor[HTML]{E0C5E9}0.583          & \cellcolor[HTML]{D8B5E3}0.507          & \cellcolor[HTML]{DEC2E8}0.612          & \cellcolor[HTML]{DFC4E9}0.397          \\
    \ding{51}                    & \ding{51}       & \ding{51}            & \cellcolor[HTML]{D5B1E1}\textbf{0.627} & \cellcolor[HTML]{D5B1E1}\textbf{0.513} & \cellcolor[HTML]{D5B1E1}\textbf{0.646} & \cellcolor[HTML]{D5B1E1}\textbf{0.412} \\ \hline
    \end{tabular}
\end{table}

\subsubsection{Effect of reconstruction modulator}

\begin{figure}[htbp]
    \centering
    \includegraphics[width=1\columnwidth]{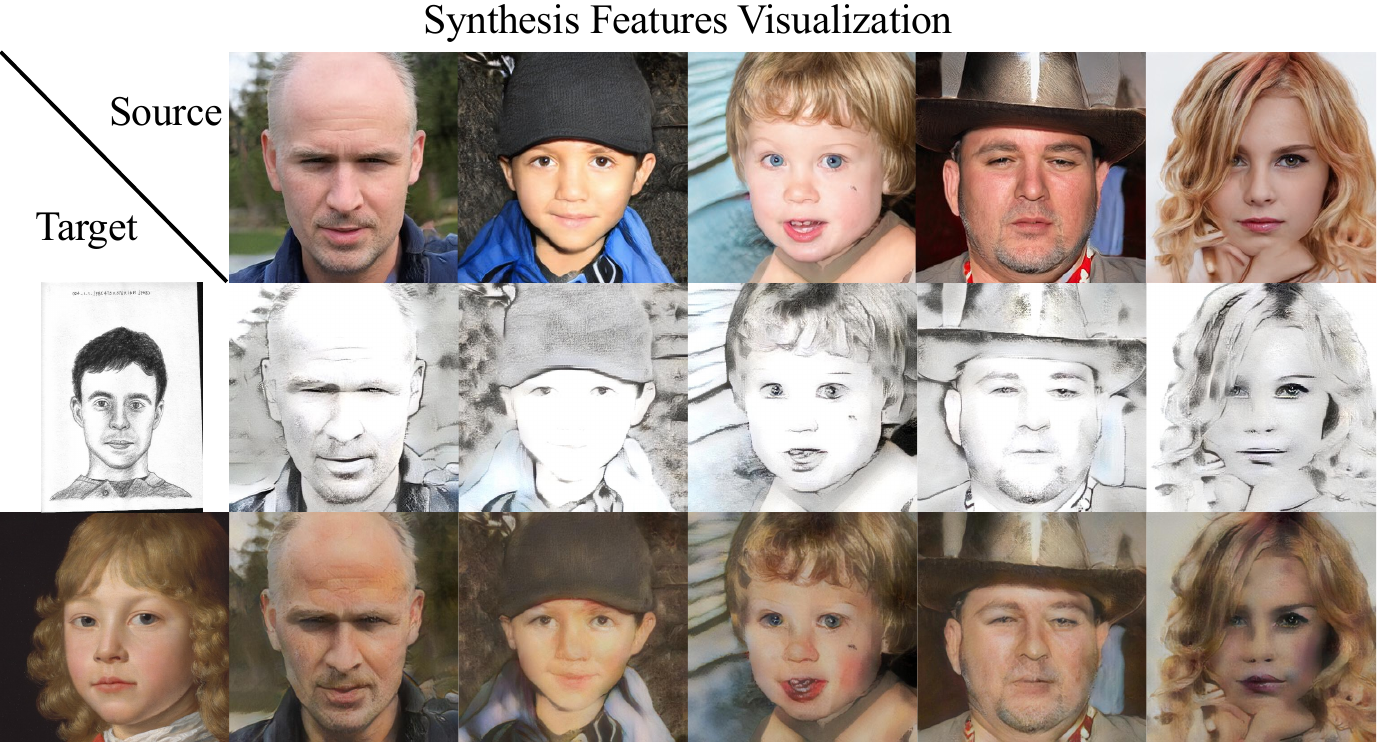} 
    \vspace{-5mm}
    \caption{Visualization experiment of synthesis feature $M_{{C_S}{S_T}}$. We use the style-content features of the source domain and the target domain to construct the synthesis feature, and then train an image decoder with the corresponding original image as the training set, and then input the synthesis feature $M_{{C_S}{S_T}}$ to obtain the visualization image.}
    \label{fig:synthesis}
\end{figure}

To validate the efficacy of the proposed style-content decoupler in feature disentanglement and the reconstruction modulator in feature synthesis, a systematic ablation study is conducted to examine: (1) whether the synthesized features retain the critical identity knowledge from the source domain, and (2) whether these reconstructed representations preserve the essential style information inherent in the target domain. 
Based on this foundation, visualization experiments should be conducted on the synthesized feature $M_{{C_S}{S_T}}$ to examine whether it preserves the source domain identity knowledge while maintaining the target domain style characteristics.
Following this framework, style-content feature pairs from source and target domains are processed through the reconstruction modulator to synthesize corresponding synthesis features $M_{{C_S}{S_S}}$ and $M_{{C_T}{S_T}}$, which are subsequently paired with their respective images $\mathbf{x}_S$ and $\mathbf{x}_T$ to establish aligned training tuples for training an image generator, thereby establishing explicit mappings between synthesis features $M$ and original images $\mathbf{x}$.
Then $M_{{C_S}{S_T}}$ is used as the input of the image generator to obtain the corresponding generated image, as shown in Fig.~\ref{fig:synthesis}.
Due to the small number of training samples, the quality of the images generated by the generator is not high, but the basic source domain identity knowledge and target domain style can be clearly seen through the reconstructed images. 
This experiments show that the synthesized features obtained by the style-content decoupler and the reconstruction modulator can effectively capture the identity knowledge of the source domain and the style information of the target domain, thus providing a solid foundation for the subsequent identity consistency constraints.

\subsubsection{The proportion of hyperparameter}

\begin{figure}[htbp]
    \centering
    \includegraphics[width=1\columnwidth]{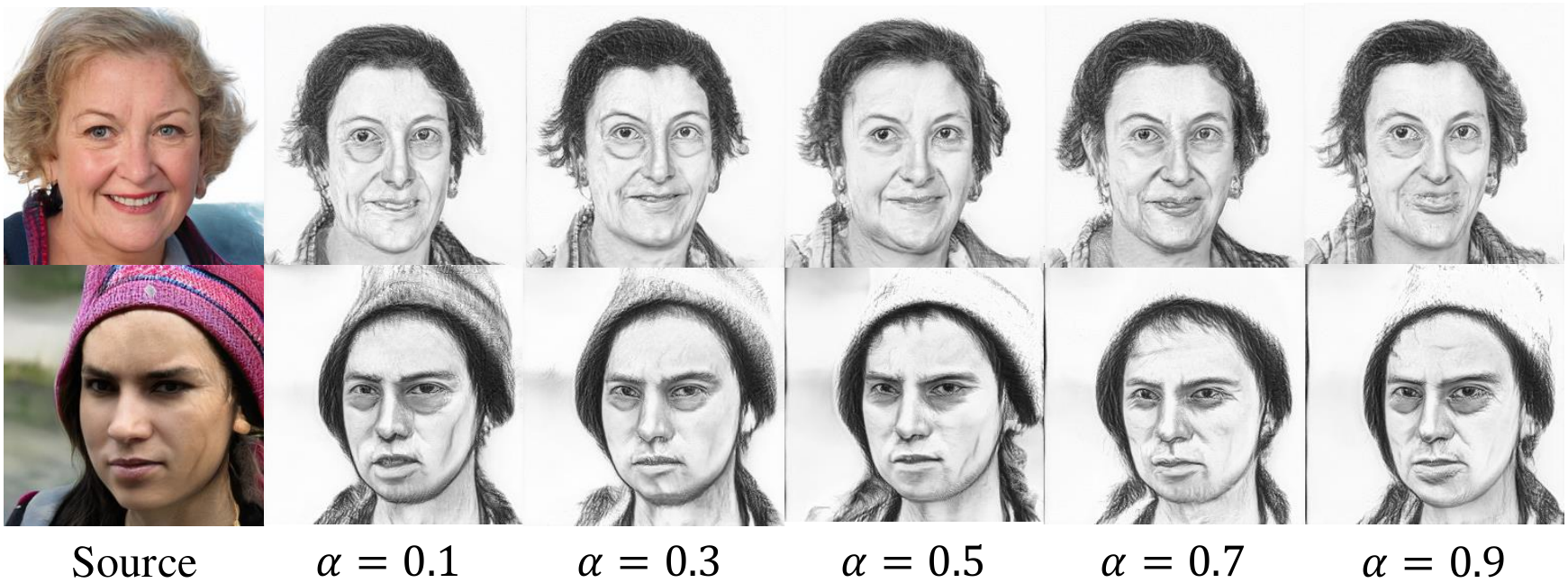} 
    \vspace{-5mm}
    \caption{Qualitative ablation of the coefficient $\alpha$ in Eq.~\ref{eq:injection}, which means the degree of identity injection}
    \label{fig:ablation-alpha}
\end{figure}

\paragraph{Hyperparameter $\alpha$ in Identity Injection}
We perform systematic parameter scanning on the identity injection weight $\alpha$ (Eq.~\ref{eq:injection}) across the interval [0.1, 0.9] with 0.2 increments, qualitatively analyzing the correlation between injection intensity and generation quality through visual inspection of target domain outputs.
As illustrated in Fig.~\ref{fig:ablation-alpha}, increasing the identity injection weight $\alpha$ from $0.1$ to $0.5$ improves the quality of the generated images, as they retain more source domain identity knowledge. 
However, when $\alpha > 0.5$, increasing values result in progressive quality degradation and perceptible artifacts attributable to oversaturation of source-domain identity features.
Thus, we therefore set $\alpha=0.5$ as the optimal identity injection coefficient.

\begin{figure*}[htbp]
    \centering
    \includegraphics[width=1\textwidth]{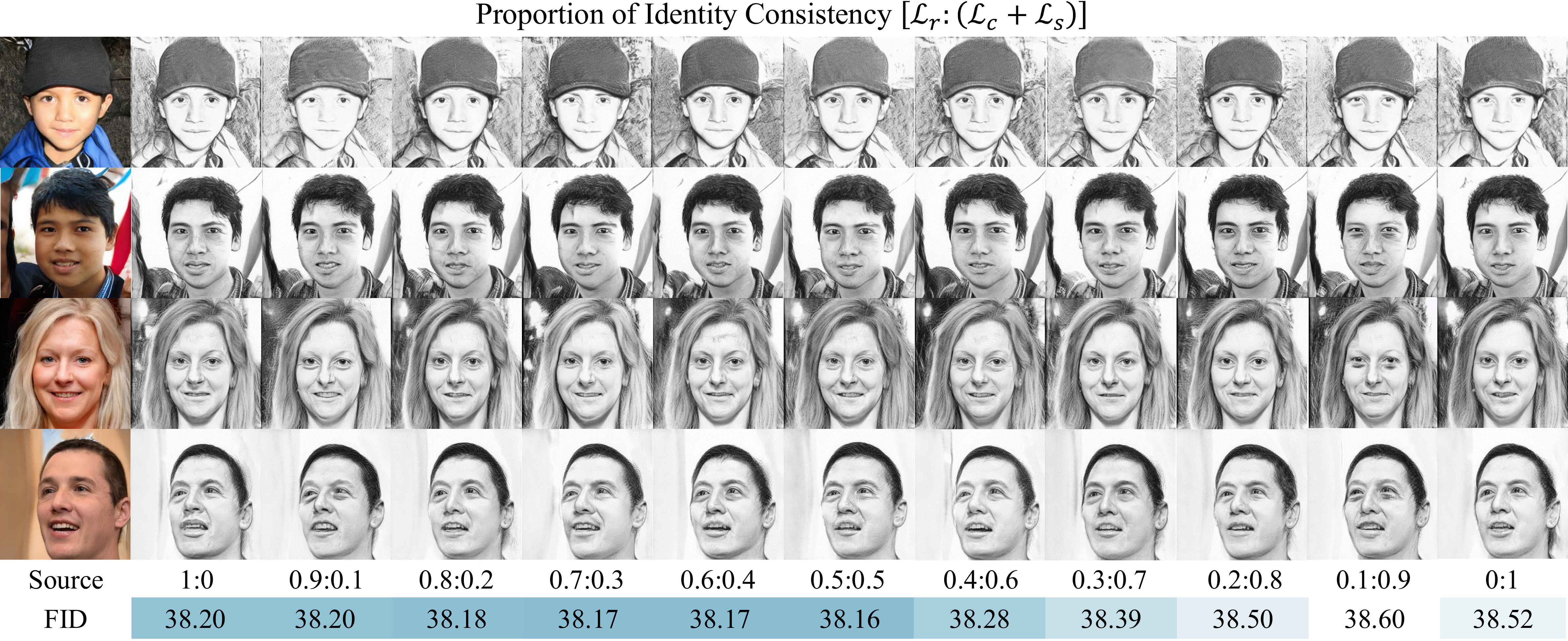}
    \vspace{-5mm}
    \caption{Qualitative and quantitative ablation experiments on the proportion of two types of losses in the Identity Preservation module: synthesis constraint and content constraint \& style constraint.}
    \label{fig:sup-ablation2}
\end{figure*}

\begin{figure}[htbp]
    \centering
    \includegraphics[width=1\columnwidth]{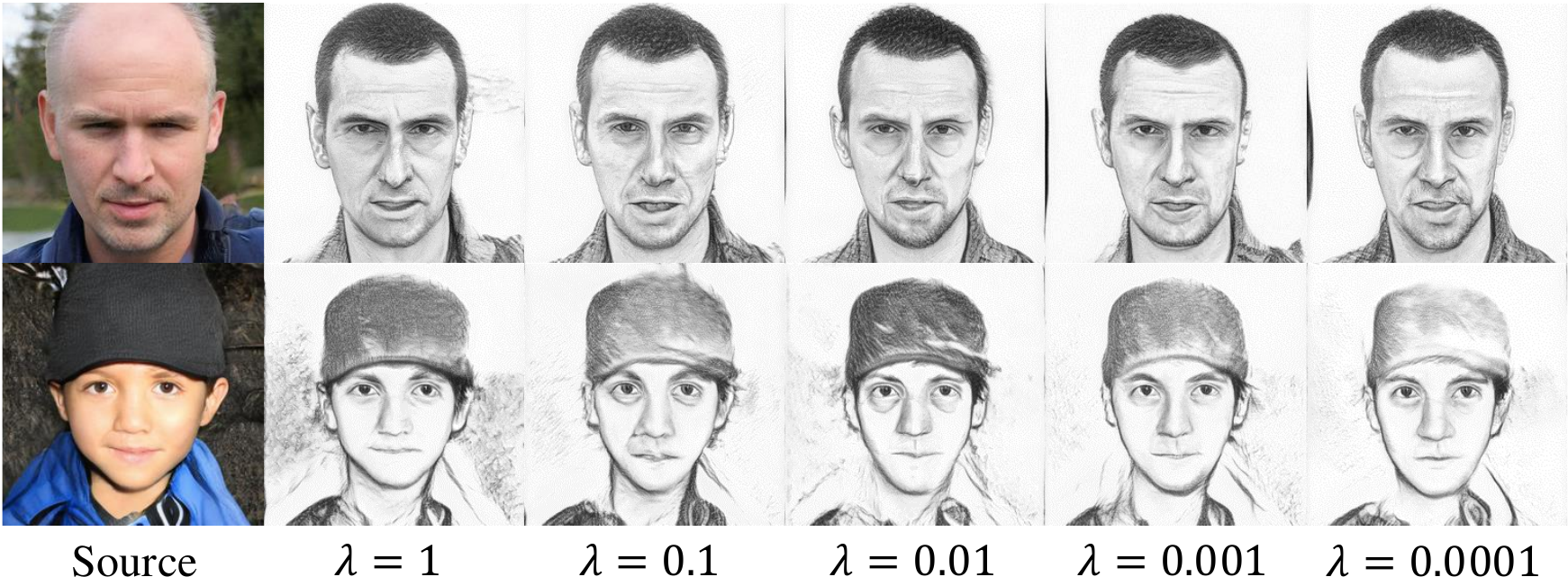} 
    \vspace{-5mm}
    \caption{Qualitative ablation of the coefficient $\lambda$ in Eq.~\ref{eq:total-loss}, which means the weights of the identity consistency}
    \label{fig:ablation-lambda}
\end{figure}

\paragraph{Hyperparameter $\lambda$ in Identity Preservation}
Additionally, we examine the impact of the hyperparameter $\lambda$ in Eq.~\ref{eq:total-loss} on the identity preservation module's performance. For the $\lambda$ ablation experiment, we conducted experiments on its magnitude, so we set the value of $\lambda$ to $1$, $1e^{-1}$, $1e^{-2}$, $1e^{-3}$ and $1e^{-4}$. As shown in Fig.~\ref{fig:ablation-lambda}, when the magnitude of $\lambda$ decreases from $1$ to $1e^{-4}$, there is a noticeable decline in both the retention of source domain identity knowledge and the overall quality of the generated images. Consequently, we set the final $\lambda$ value to $1$ to ensure optimal performance in our method.

\paragraph{The proportion of \texorpdfstring{$\mathcal{L}_r$}{Lr} and \texorpdfstring{$\mathcal{L}_c\&\mathcal{L}_s$}{Lc\&Ls}}

We conduct systematic experiments to analyze the weighting scheme between synthesis constraint ($\mathcal{L}_r$) and content-style constraints ($\mathcal{L}_c \& \mathcal{L}_s$), revealing their joint effects on both the perceptual quality and quantitative metrics of target domain generations.
The quantitative impact of constraint weighting ratios is systematically analyzed with visual corroboration provided in Fig.~\ref{fig:sup-ablation2}.
Here we take the source domain generative model on FFHQ and the target domain few-shot data as sketches.
First, we focus on the quantitative experiment of the proportion of the $\mathcal{L}_r$ and $\mathcal{L}_c \& \mathcal{L}_s$, as shown in Fig.~\ref{fig:sup-ablation2}. 
We can clearly observe that the FID score monotonically decreases as the $\mathcal{L}_r$ to $\mathcal{L}_c\&\mathcal{L}_s$ ratio approaches equilibrium at $0.5:0.5$, achieving minimal value $(38.16)$ that quantitatively demonstrates optimal generation quality. 
In addition, through qualitative experimental analysis, as shown in Fig.~\ref{fig:sup-ablation2}. 
We can observe that the proportions at $1:0$ or $0:1$ also represent the extreme cases in the Sec~\ref{sec4:effect-optimization}, that is, only $\mathcal{L}_r$ plays a role and only $\mathcal{L}_c \& \mathcal{L}_s$ plays a role, which also have been described the loss of identity in the previous section.
As the proportion of $\mathcal{L}_r$ decreases  and the proportion of $\mathcal{L}_c \& \mathcal{L}_s$ increases, the source domain content knowledge of the target domain generated image gradually increases, and the identity knowledge gradually becomes similar to the source domain generated image. 
However, when the proportion is smaller than $0.5:0.5$, $\mathcal{L}_c \& \mathcal{L}_s$ gradually becomes dominant, and the image gradually begins to become distorted and overly sharp.
This also shows the effectiveness of our identity consistency construction.

\subsubsection{Computational efficiency}

\begin{table}[htbp]
    \normalsize
    \centering
    \caption{Computational complexity experiment with baseline}
    \begin{tabular}{cccc|c}
    \hline
                & CDC      & RSSA     & PIR      & I\textsuperscript{2}P \\ \hline
    Time(min)   & 80   & 103  & 130  & 88   \\
    Model size(MB) & 115.28 & 115.28 & 115.28 & 115.28  \\
    Memory size(GB) & 12.9 & 18.9 & 19.2  & 14.7   \\ \hline
    \end{tabular}
    \label{tab:computation}
\end{table}

As shown in Table~\ref{tab:computation}, comparative test of the computational efficiency of our I\textsuperscript{2}P with several baseline methods. 
The batch size of the test is 4, and the experiments are all conducted on a single NVIDIA 3090 with 24GB of memory and 5002 steps.
Both our I\textsuperscript{2}P and baseline methods adopt the StyleGANv2~\cite{karras2020analyzing} architecture, resulting in equivalent parameter scales for the final translated generator $G_T$. 
Additionally, whereas RSSA~\cite{xiao2022few} requires spatial mapping networks and PIR~\cite{heFewshotImageGeneration2023a} incorporates image translation modules, our framework achieves enhanced computational efficiency in both time complexity and memory footprint.

\section{Conclusion and Limitations}
\label{sec5:conclusion}

In this paper, we introduce the \textbf{I}dentity \textbf{I}njection and \textbf{P}reservation~(\textbf{I\textsuperscript{2}P}) method for few-shot generative model adaptation.
By incorporating identity-preserving transformations during pretraining, I\textsuperscript{2}P effectively transfers identity-related features from the source domain to the target domain, preserving crucial identity knowledge even with limited data.
Our extensive experiments show that I\textsuperscript{2}P outperforms previous techniques, achieving superior results on small datasets, which enhances the model's robustness and flexibility.
However, the method's effectiveness depends on the quality of identity-preserving transformations and requires careful hyperparameter tuning. In domains with abstract features or inconsistent identity concepts (e.g., Human $\rightarrow$ Cats, Human $\rightarrow$ Dogs), I\textsuperscript{2}P may be less effective.
Despite these limitations, further research will refine our approach and advance data-efficient generative models.
Future work may also explore automated transformation selection and broader applications across more diverse generative tasks, potentially improving generalization in real-world, low-resource scenarios.

{
    \bibliographystyle{unsrt}
    \bibliography{main}
}

\clearpage
\newpage
\begin{IEEEbiographynophoto}{Yeqi He}
received the B.E. degree in intelligence science and technology from Hangzhou Dianzi University, Hangzhou, China, in 2023.
He is currently pursuing the Ph.D. degree with Hangzhou Dianzi University, Hangzhou, China.
His research interests include computer vision and generative models.
\end{IEEEbiographynophoto}

\begin{IEEEbiographynophoto}{Liang Li} 
received his B.S. degree from Xi'an Jiaotong University in 2008, and Ph.D. degree from Institute of Computing Technology, Chinese Academy of Sciences, Beijing, China in 2013. 
From 2013 to 2015, he held a post-doc position with the Department of Computer and Control Engineering, University of Chinese Academy of Sciences, Beijing, China. 
Currently he is serving as the associate professor at Institute of Computing Technology, Chinese Academy of Sciences. 
He has also served on a number of committees of international journals and conferences. 
Dr. Li has published over 100 refereed journal/conference papers, including IEEE TPAMI, IJCV, TNNLS, TIP, CVPR, ICCV.
His research interests include multimedia content analysis, computer vision, and pattern recognition.
\end{IEEEbiographynophoto}

\begin{IEEEbiographynophoto}{Jiehua Zhang} 
received the B.E. degree from Hunan Institute of Engineering, Xiangtan, China, in 2019, and M.S. degree from Hangzhou Dianzi University, Hangzhou, China, in 2022. 
He is currently pursuing the Ph.D. degree with Xi'an Jiaotong University, Xi'an, China. 
His main research interests include computer vision and transfer learning.
\end{IEEEbiographynophoto}

\begin{IEEEbiographynophoto}{Yaoqi Sun}
received the BS degree from  Hangzhou Dianzi University, Hangzhou, China, in 2016.
He is currently working toward the PHD degree at  Hangzhou Dianzi University. 
His major research interest includes computer vision problems related to transfer learning. He has published over 20 refereed journal/conference papers, including IEEE TPAMI, TCSVT, ACM TOMM, ACM MM, etc.
\end{IEEEbiographynophoto}

\begin{IEEEbiographynophoto}{Xichun Sheng}
received the M.S. degree in Materials Science and Engineering from Xiangtan University. 
Currently pursuing the Ph.D. degree in Computer Technology at Macau Polytechnic University. 
Researching interests include visual language models, which integrate computer vision and natural language processing to enhance the capabilities of artificial intelligence systems in understanding and generating multimodal content.
\end{IEEEbiographynophoto}

\begin{IEEEbiographynophoto}{Zhidong Zhao}
(Member, IEEE) received the B.S. and M.S. degrees in mechanical engineering from the Nanjing University of Science and Technology, Nanjing, China, in 1998 and 2001, respectively, and the Ph.D. degree in biomedical engineering from Zhejiang University, Hangzhou, China, in 2004. 
He is currently a Full Professor with Hangzhou Dianzi University, Hangzhou. 
His research interests include biomedical signal processing, wireless sensor networks, biometrics, and machine learning.
\end{IEEEbiographynophoto}

\begin{IEEEbiographynophoto}{Chenggang Yan}
received the B.S. degree in control science and engineering from Shandong University, Jinan, China, in 2008, and the Ph.D. degree in computer science from the Chinese Academy of Sciences University, Beijing, China, in 2013. 
He is currently a Professor with the Department of Automation, Hangzhou Dianzi University, Hangzhou, China. 
His research interests include computational photography, pattern recognition, and intelligent systems.
\end{IEEEbiographynophoto}

\end{document}